\documentclass[runningheads]{llncs}
\usepackage{graphicx}

\usepackage{tikz}
\usepackage{comment}
\usepackage{amsmath,amssymb} %
\usepackage{color}
\usepackage[font=small]{caption}

\usepackage[pagebackref,breaklinks,colorlinks]{hyperref}
\usepackage{booktabs}
\usepackage{multirow}
\usepackage[official]{eurosym}
\usepackage{mathrsfs}
\usepackage{yfonts}
\usepackage{xspace}
\usepackage{pifont} %
\usepackage{enumitem}

\newcommand{\method}{Text-to-Motions\xspace}
\newcommand{\methodshort}{\texttt{TEMOS\xspace}}
\definecolor{aliceblue}{rgb}{0.94, 0.97, 1.0}

\def\sepappendix{1}

\newcommand{\cmark}{\ding{51}}%
\newcommand{\xmark}{\ding{55}}%

\def\sepappendix{0}

\begin{document}
\pagestyle{headings}
\mainmatter

\title{\methodshort: Generating diverse human motions from textual descriptions}

\titlerunning{\methodshort}

\author{Mathis Petrovich\inst{1,2} \and Michael J. Black\inst{2} \and G\"ul Varol\inst{1}
}
\institute{
LIGM, \'Ecole des Ponts, Univ Gustave Eiffel, CNRS, France \and
Max Planck Institute for Intelligent Systems, T\"{u}bingen, Germany \\
\email{\{mathis.petrovich,gul.varol\}@enpc.fr, black@tue.mpg.de}\\
\url{https://mathis.petrovich.fr/temos/}
}
\authorrunning{M.~Petrovich et al.}
\maketitle

\begin{abstract}
We address the problem of generating diverse 3D human motions from textual descriptions. This challenging task requires joint modeling of both modalities: understanding and extracting useful human-centric information from the text, and then generating plausible and realistic sequences of human poses. 
In contrast to most previous work which focuses on generating a single, deterministic, motion from a textual description,
we design a variational approach that can produce \textit{multiple} diverse human motions. We propose \methodshort, a text-conditioned generative model leveraging variational autoencoder (VAE) training with human motion data, in combination with a text encoder that produces distribution parameters compatible with the VAE latent space. We show the \methodshort{} framework can produce both skeleton-based animations as in prior work, as well more expressive SMPL body motions. 
We evaluate our approach on the KIT Motion-Language benchmark and, despite being relatively straightforward, demonstrate significant improvements over the state of the art. Code and models are available on our webpage.

\end{abstract}
\section{Introduction}
\label{sec:intro}

We explore the problem of generating 3D human motions, i.e., sequences of 3D poses, from natural language textual descriptions (in English in this paper). 
Generating text-conditioned human motions has numerous applications
both for the virtual (e.g., game industry) and real worlds (e.g., controlling a robot with speech for personal physical assistance). For example,
in the film and game industries, motion capture is often used to create special effects featuring humans.
Motion capture is expensive, therefore
technologies that automatically synthesize new motion data could save time and money.

Language represents a natural interface for people to interact 
with computers~\cite{hill1983}, and our work
provides a foundational step towards creating human animations using natural 
language input.
The problem of generating human motion from free-form text, however, is relatively new since it relies on advances in both language modeling and human motion synthesis.
Regarding the former, we build on advances in language modeling using transformers.
In terms of human motion synthesis, much of the previous work has focused on generating motions conditioned on a single action label, not a sentence, e.g.,~\cite{chuan2020action2motion,petrovich21actor}.
Here we go further by encoding both the language and the motion using transformers in a joint latent space.
The approach is relatively straightforward, yet achieves results that significantly outperform the latest state of the art.
We perform extensive experiments and ablation studies to understand which design choices are critical.

\begin{figure}[t]
    \centering
    \includegraphics[width=.99\linewidth]{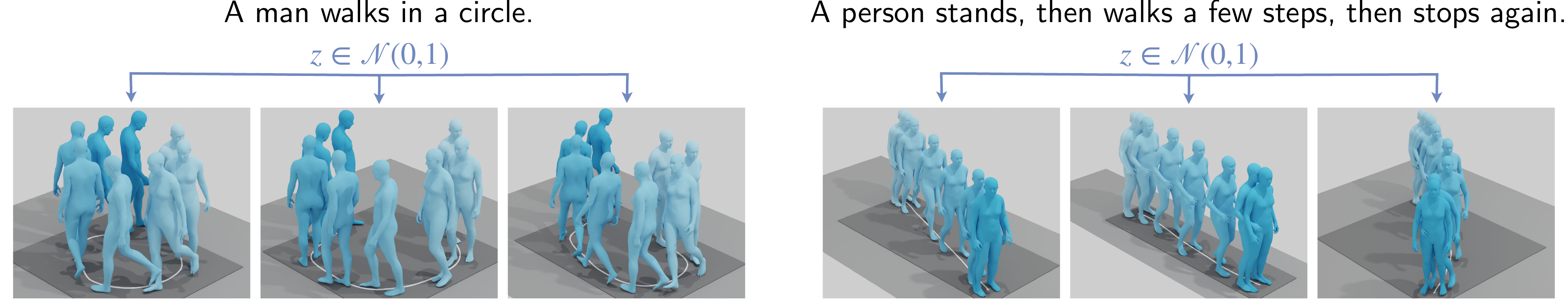}
    \caption{\textbf{Goal:}
        \method (\methodshort)
        learns to synthesize
        human motion sequences
        conditioned on a textual description and a duration.
        SMPL pose sequences are generated by sampling from a single latent vector, $z$.
        Here, we illustrate the diversity of our motions on two sample texts,
        providing three generations per text input.
        Each image corresponds to a motion sequence where we visualize
        the root trajectory projected on the ground plane and the human poses at multiple equidistant
        time frames. The flow of time is shown with a color code where lighter blue denotes the past.
    }
    \label{fig:teaser}
\end{figure}

Despite recent efforts in this area, most current methods  generate only \textit{one} output motion per text input~\cite{lin2018,Ahuja2019Language2PoseNL,Ghosh_2021_ICCV}. That is, with the input ``A man walks in a circle'', these methods synthesize one motion. However,
one description often can map to \textit{multiple} ways of performing the actions,
often due to ambiguities and lack of details, 
e.g., in our example, the size and the orientation of the circle are not specified.
An ideal generative model should therefore be able to synthesize multiple
sequences that respect the textual description while exploring the degrees
of freedom to generate natural variations.
While, in theory, the more precise the description becomes, the less
space there is for diversity; it is a desirable property
for natural language interfaces to manage intrinsic ambiguities of linguistic expressions~\cite{Gao2015}.
In this paper, we propose a method that allows sampling from a distribution
of human motions conditioned on natural language descriptions. 
Figure~\ref{fig:teaser} illustrates multiple sampled
motions generated from two input texts; check the project webpage~\cite{projectpage_temos} for video examples.

A key challenge is building models that are effective
for temporal modeling. 
Most prior work employs autoregressive models
that iteratively decode the next time frame given the past.
These approaches may suffer from drift over time and often, eventually, produce static poses \cite{Martinez_2017_CVPR}.
In contrast, sequence-level generative models encode an entire sequence and can exploit long-range context.
In this work, we incorporate
the powerful
Transformer models \cite{vaswani2017attention}, which have proven effective for various
sequence modeling tasks~\cite{devlin2018bert,bain21_frozen}.
We design a simple yet effective architecture, where both the 
motion and text are input to Transformer encoders before
projecting them to a cross-modal joint space.  
Similarly, the motion
decoder uses a Transformer architecture
taking positional encodings and a latent vector as input, and generating a 3D human motion (see Figure~\ref{fig:model}).
Notably, %
a single sequence-level latent vector is used to
decode the motion in one shot, without any autoregressive
iterations. 
Through detailed ablation studies, %
we show that the main
improvement over prior work stems from this design.

A well-known challenge common to generative models
is the difficulty of evaluation. 
While many metrics are used in evaluating generated motions, each of them is limited.
Consequently, in this work,
we rely on both quantitative measures that compare
against the ground truth motion data associated with
each test description, and human perceptual studies
to evaluate the perceived quality of the motions.
The former is problematic particularly for this
work, because it assumes one true motion per text,
but our method produces multiple motions due to its
probabilistic nature.
We find that human judgment of motion quality is necessary for a full picture.

Moreover, the state of the art reports 
results on the task of future motion prediction.
Specifically, Ghosh et al.~\cite{Ghosh_2021_ICCV}
assume the first pose in the generated sequence is available from the ground truth.
In contrast, we evaluate our method by synthesizing
the full motion from scratch; i.e.~without conditioning on the first, ground truth, frame. 
We provide results
for various settings, e.g., comparing a random 
generation against the ground truth, or picking
the best out of several generations.
We outperform previous work even when sampling
a single random generation, but the performance
improves as we increase the number of generations
and pick the best.

A further addition we make over existing text-to-motion
approaches is to generate sequences of SMPL
body models~\cite{smpl2015}.
Unlike classical skeleton representations, the parametric
SMPL model provides the body surface, which can support future research on motions that involve interaction with objects or the scene.
Such skinned generations were considered in other
work on unconstrained or action-conditioned
generation \cite{Zhang2020WeAM,petrovich21actor}.
Here, we demonstrate promising results for the text-conditioning
scenario as well.
The fact that the framework supports multiple body representations, illustrates its generality.

In summary, our contributions are the following:
(i) We present \method (\methodshort), a novel
cross-modal variational model that can produce diverse 3D human movements given textual descriptions in natural language.
(ii)~In our experiments, we provide an extensive ablation study of the model
components and outperform the state of the art by a large margin both on standard metrics
and through perceptual studies. %
(iii) We go beyond stick figure generations, and exploit the SMPL model for
text-conditioned body surface synthesis, demonstrating qualitatively
appealing results.
The code and trained models are available on our project page~\cite{projectpage_temos}.

\section{Related work}
We provide a brief summary of relevant work on
human motion synthesis and text-conditioned motion generation.
While there is also work on facial motion generation~\cite{Karras2017,VOCA2019,Richard_2021_ICCV,Fan_2022_CVPR}, here we focus on articulated human bodies.

\noindent\textbf{Human motion synthesis.} While there is a 
large body of work focusing on future human motion 
prediction \cite{Pavllo2018QuaterNetAQ,Habibie2017ARV,BarsoumCVPRW2018,Aksan_2019_ICCV,Zhang2020WeAM,yuan2020dlow}
and completion \cite{duan2021singleshot,Harvey:ToG:2020},
here, we give an overview of
methods that generate motions from scratch (i.e., no past or future observations).
Generative models of human motion have been designed using GANs
\cite{Ahn2018Text2ActionGA,Lin2018HumanMM},
VAEs \cite{chuan2020action2motion,petrovich21actor}, or normalizing flows
\cite{Henter2020,weaklynormalizeflowZanfir2020}.
In this work, we employ VAEs in the context of Transformer neural network architectures.
Recent work suggest that VAEs are effective for human motion generation compared with GANs~\cite{chuan2020action2motion,petrovich21actor},
while being easier to train.

Motion
synthesis methods can be broadly divided into two categories:
(i) unconstrained generation, which models the entire space of possible motions 
\cite{Yan_2019_ICCV,Zhao2020BayesianAH,Zhang2020PerpetualMG} and (ii) 
conditioned synthesis, which aims for controllability such as using
music \cite{dancing2music2019,Li2020LearningTG,li2021learn},
speech \cite{bhattacharaya:2021,ginosar2019gestures},
action \cite{chuan2020action2motion,petrovich21actor}, and
text \cite{lin2018,Lin2018HumanMM,Ahn2018Text2ActionGA,Ahuja2019Language2PoseNL,Ghosh_2021_ICCV,saunders2021mixed} conditioning.
Generative models that synthesize unconstrained motions
aim, by design, to sample from a distribution,
allowing generation of diverse motions.
However, they lack the ability to control the generation process.
On the other hand, the conditioned synthesis can be further divided
into two categories:
deterministic \cite{lin2018,Ahuja2019Language2PoseNL,Ghosh_2021_ICCV} or
probabilistic \cite{Ahn2018Text2ActionGA,Lin2018HumanMM,dancing2music2019,Li2020LearningTG,chuan2020action2motion,petrovich21actor}.
In this work, we focus on the latter, motivated by the fact that
there are often multiple possible motions for a given condition.

\noindent\textbf{Text-conditioned motion generation.} 
Recent work explores the advances in natural language modeling 
\cite{NIPS2013_9aa42b31,devlin2018bert}
to design sequence-to-sequence approaches to cast
the text-to-motion task as a machine translation problem
\cite{Plappert2018LearningAB,lin2018,Ahn2018Text2ActionGA}.
Others build joint cross-modal embeddings to map the text
and motion to the same space \cite{Yamada2018PairedRA,Ahuja2019Language2PoseNL,Ghosh_2021_ICCV}, which has been a success in other research area~\cite{clip-pmlr-v139-radford21a,liu2022florenceanew,Yang_2022_CVPR,frozen-Bain21}.

Several methods use an impoverished body motion representation.
For example, some do not model the global trajectory
\cite{Plappert2018LearningAB,Yamada2018PairedRA}, making
the motions unrealistic and ignoring %
the global movement description
in the input text. 
Text2Action~\cite{Ahn2018Text2ActionGA} uses a sequence-to-sequence model but only models the upper body motion.
This is because Text2Action uses a semi-automatic approach to create training data from the MSR-VTT captioned video dataset~\cite{Xu2016msrvtt}, which contains frequently occluded lower bodies. 
They apply 2D pose estimation, lift the joints to 3D, and employ manual cleaning of the input text to make it generic.

Most other work uses 3D motion capture data 
\cite{lin2018,Lin2018HumanMM,Ahuja2019Language2PoseNL,Ghosh_2021_ICCV}.
DVGANs~\cite{Lin2018HumanMM} adapt the CMU MoCap database~\cite{cmuWEB} and Human3.6M~\cite{IonescuSminchisescu11,h36m_pami}
for the task of motion generation and completion, and they use
the action labels as text-conditioning instead of categorical supervision.
More recent works %
\cite{lin2018,Ahuja2019Language2PoseNL,Ghosh_2021_ICCV}
employ the KIT Motion-Language %
dataset
\cite{Plappert2016_KIT_ML}, which is also the focus of our work.

A key limitation of many state-of-the-art text-conditioned
motion generation models is that they are deterministic \cite{Ahuja2019Language2PoseNL,Ghosh_2021_ICCV}.
These methods employ a shared cross-modal latent space approach.
Ahuja et al.~\cite{Ahuja2019Language2PoseNL} employ word2vec text embeddings \cite{NIPS2013_9aa42b31},
while \cite{Ghosh_2021_ICCV} uses the more recent BERT model \cite{devlin2018bert}.

Most similar to our work is Ghosh~et.~al.~\cite{Ghosh_2021_ICCV}, which builds on
Language2Pose \cite{Ahuja2019Language2PoseNL}. 
Our key difference is the
integration of a variational approach for sampling a diverse set of motions from a single text. 
Our further improvements include the use of Transformers to encode motion sequences into a single embedding instead of the autoregressive approach in \cite{Ghosh_2021_ICCV}.
This allows us to encode distribution parameters of the VAE as in~\cite{petrovich21actor},
proving effective in our state-of-the-art results.
Ghosh et al.~\cite{Ghosh_2021_ICCV} also encode the upper body and lower body
separately, whereas our approach does not need such hand crafting.

\section{Generating multiple motions from a textual description}
\label{sec:method}

\begin{figure}[t]
    \centering
    \includegraphics[width=.9\linewidth]{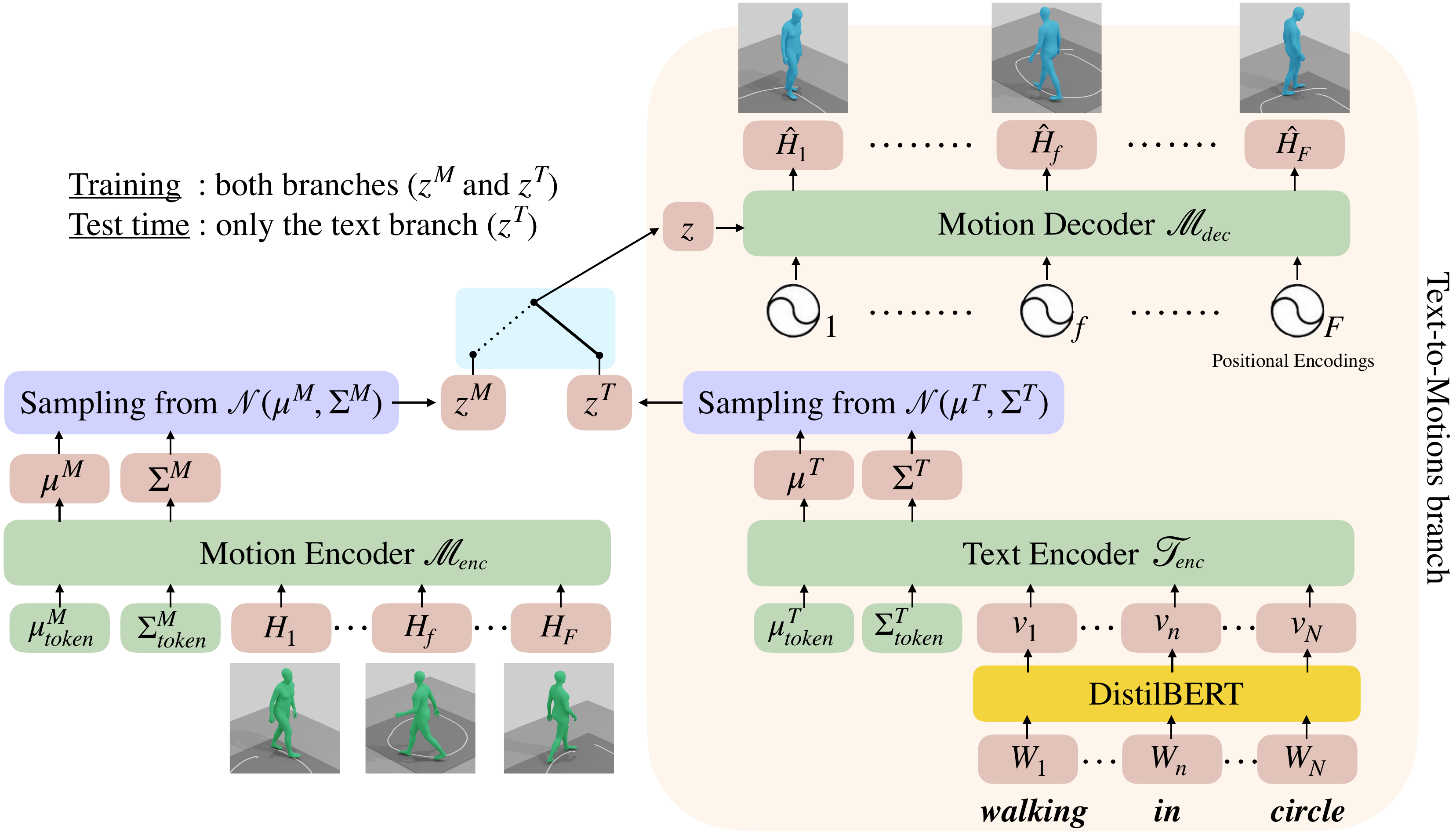}
    \caption{\textbf{Method overview:}
        During training, we encode both the motion and text through their respective Transformer encoders,
        together with modal-specific learnable distribution tokens. The encoder outputs corresponding to these
        tokens provide Gaussian distribution parameters on which the KL losses are applied and a latent vector $z$ is sampled.
        Reconstruction losses
        on the motion decoder outputs further provide supervision for both motion and text branches.
        In practice, our word embedding consists of a variational encoder that takes input from a pre-trained and frozen DistilBERT~\cite{distilbert_sanh} model.
        Trainable layers are denoted in green, the inputs/outputs in brown.
        At test time, we only use the right branch, which goes from an input text to
        a diverse set of motions through the random sampling of the latent vector $z^T$ on the cross-modal space.
        The output motion duration %
        is determined 
        by the number of positional encodings $F$.
    }
    \label{fig:model}
\end{figure}

In this section, we start by formulating the problem (Section~\ref{subsec:task}).
We then provide details on our model design (Section~\ref{subsec:envae}),
as well as our training strategy (Section~\ref{subsec:training}).

\subsection{Task definition}
\label{subsec:task}

Given a sentence describing a motion, the goal is to generate various 
sequences of 3D human poses and trajectories that match the textual input.
Next, we describe the representation for the text and motion data.

\noindent\textbf{Textual description} represents a written natural language sentence (e.g., in 
English) that describes what and how a human motion is performed. The sentence
can include various levels of detail:
a precise sequence of actions such as \textit{``A human walks two steps and then stops''} or
a more ambiguous description such as \textit{``A man walks in a circle''}.
The data structure is a sequence of words
$W_{1:N} = W_1, \ldots, W_N$ from the English vocabulary.

\noindent\textbf{3D human motion} is defined as a sequence of 
human poses $H_{1:F} = H_1, \ldots, H_F$,
where $F$ denotes the number of time frames. 
Each pose $H_f$ corresponds to a representation of the articulated human body.
In this work, we employ two types of body motion representations:
one based on skeletons, one based on SMPL~\cite{smpl2015}.
First, to enable a comparison with the state of the art, we follow
the rotation-invariant skeleton representation from Holden~et.~al.~\cite{holden2016motionsynthesis},
which is used in the previous work we compare with 
\cite{Ahuja2019Language2PoseNL,Ghosh_2021_ICCV}. 
Second, we incorporate the parametric SMPL representation by encoding the global root trajectory of the body and parent-relative joint rotations in 6D representation~\cite{Zhou2019OnTC}. We provide
detailed formulations for both motion representations in \if\sepappendix1{Appendix~B.}
\else{Appendix~\ref{app:sec:representation}.}
\fi

More generally, a human motion can be represented by a sequence of $F$ poses
each with $p$ dimensions,
so that at frame $f$, we have $H_f \in \mathbb{R}^{p}$.
Our goal is, given a textual description $W_{1:N}$, to sample
from a distribution of plausible motions $H_{1:F}$ and to generate
multiple hypotheses.

\subsection{TEMOS model architecture}
\label{subsec:envae}
Following \cite{Ahuja2019Language2PoseNL}, we learn
a joint latent space between the two modalities: motion and text %
(see Figure~\ref{fig:model}).
To incorporate generative modeling in such an approach,
we employ a VAE~\cite{kingma2014auto} formulation that requires architectural changes.
We further employ Transformers~\cite{vaswani2017attention}
to obtain sequence-level embeddings both for the text and motion data.
Next, we describe the two encoders for motion and text, 
followed by the motion decoder.

\noindent\textbf{Motion and text encoders.}
We have two encoders for representing
motion $\mathscr{M}_{enc}$ and text $\mathscr{T}_{enc}$
in a joint space.
The encoders are designed to be as symmetric
as possible across the two modalities.
To this end, we adapt the ACTOR~\cite{petrovich21actor} Transformer-based VAE motion encoder
by making it class-agnostic
(i.e., removing action conditioning). This encoder takes as input a sequence
of vectors of arbitrary length, as well as learnable \textit{distribution tokens}. %
The outputs
corresponding to the distribution tokens are treated as
Gaussian distribution parameters $\mu$ and $\Sigma$ of the sequence-level latent space.
Using the reparameterization trick~\cite{kingma2014auto}, we sample  a latent vector $z \in \mathbb{R}^d$ from this distribution
(see Figure~\ref{fig:model}). The latent space dimensionality $d$ is set to 256 in our experiments.

For the motion encoder $\mathscr{M}_{enc}$, the input sequence of vectors
is $H_{1:F}$, representing the poses.
For the text encoder $\mathscr{T}_{enc}$, the inputs are word embeddings for $W_{1:N}$
obtained from a pretrained language model DistilBERT~\cite{distilbert_sanh}.
We freeze the weights of DistilBERT unless stated otherwise.

\noindent\textbf{Motion decoder.} The motion decoder $\mathscr{M}_{dec}$ is
a Transformer decoder
(as in ACTOR~\cite{petrovich21actor}, but without the bias token to make it class agnostic),
so that given a latent vector $z$ and a duration $F$,
we generate a 3D human motion sequence $\widehat{H}_{1:F}$ non-autoregressively from a single latent vector.
Note that such approach does not require masks in self-attention,
and tends to provide a globally consistent motion. %
The latent vector is obtained from one of the two encoders during training (described next, in Section~\ref{subsec:training}),
and the duration is represented as a sequence of positional encodings in the form of sinusoidal functions.
We note that our model can produce variable durations, which is another source of
diversity (see supplementary video~\cite{projectpage_temos}).

\subsection{Training strategy}
\label{subsec:training}

For our cross-modal neural network training, we sample a batch of
text-motion pairs at each training iteration.
In summary, both input modalities go through their respective
encoders, and both encoded vectors go through the motion decoder
to reconstruct the 3D poses. This means we have one branch that is text-to-motion and another branch that is an autoencoder for motion-to-motion (see Figure~\ref{fig:model}).
At test time, we only use the text-to-motion branch.
This approach proved effective in previous work \cite{Ahuja2019Language2PoseNL}.
Here, we first briefly describe the loss terms to train this model \textit{probabilistically}.
Then, we provide implementation details.

Given a ground-truth pair consisting of human motion $H_{1:F}$ and textual description $W_{1:N}$, we use
(i)~two reconstruction losses -- one per modality, 
(ii)~KL divergence losses comparing each modality against Gaussion priors,
(iii)~KL divergence losses, as well as a cross-modal embedding similarity loss
to compare the two modalities to each other.

\noindent\textbf{Reconstruction losses ($\mathcal{L}_{\text{R}}$).} 
We obtain $\widehat{H}^{M}_{1:F}$ and $\widehat{H}^{T}_{1:F}$
by inputting the motion embedding and text embedding
to the decoder, respectively.
We compare these motion reconstructions to the ground-truth
human motion $H_{1:F}$ via:
\begin{equation} \label{eq:lr}
\mathcal{L}_{\text{R}} = \mathscr{L}_1(H_{1:F}, \widehat{H}^{M}_{1:F}) + \mathscr{L}_1(H_{1:F}, \widehat{H}^{T}_{1:F}) 
\end{equation}
where $\mathscr{L}_1$ denotes the smooth L1 loss.

\noindent\textbf{KL losses ($\mathcal{L}_{\text{KL}}$).} To enforce the two modalities to be close
to each other
in the latent space, we minimize the Kullback-Leibler (KL) divergences between the distributions
of the text embedding $\phi^T = \mathcal{N}(\mu^T, \Sigma^T)$ 
and the motion embedding $\phi^M = \mathcal{N}(\mu^M, \Sigma^M)$. 
To regularize the shared latent
space, we encourage each distribution to be similar to a normal distribution $\psi = \mathcal{N}(0, I)$ 
(as in standard VAE formulations).
Thus we obtain four terms:
\begin{equation}
\begin{split} \label{eq:kl}
    \mathcal{L}_{\text{KL}} &= \text{KL}(\phi^T, \phi^M) + \text{KL}(\phi^M, \phi^T) \\
    &+ \text{KL}(\phi^T, \psi) + \text{KL}(\phi^M, \psi).
\end{split}
\end{equation}

\noindent\textbf{Cross-modal embedding similarity loss ($\mathcal{L}_{\text{E}}$).} After sampling the text embedding $z^T \sim \mathcal{N}(\mu^T, \Sigma^T)$ and the motion embedding $z^M \sim \mathcal{N}(\mu^M, \Sigma^M)$ from the two encoders, we also constrain 
them to be as close as possible to each other, with the following loss term
(i.e., loss between the cross-modal embeddings): 
\begin{equation}
\begin{split} \label{eq:manifold}
    \mathcal{L}_{\text{E}} &= \mathscr{L}_1(z^T, z^M).
\end{split}
\end{equation}

The resulting total loss is defined as a weighted sum of the three terms:
$\mathcal{L} = \mathcal{L}_{\text{R}} + \lambda_{\text{KL}}\mathcal{L}_{\text{KL}} + \lambda_{\text{E}}\mathcal{L}_{\text{E}}$.
We empirically set $\lambda_{\text{KL}}$ and $\lambda_{\text{E}}$ to $10^{-5}$, and provide
ablations. While some of the loss terms may appear redundant, we experimentally
validate each term. \\

\noindent\textbf{Implementation details.}
We train our models for 1000 epochs with the AdamW
optimizer \cite{Kingma2015Adam,adamw2019} using a fixed learning rate of $10^{-4}$.
Our minibatch size is set to 32.
Our Transformer encoders and decoders consist of 6 layers for both
motion and text encoders, as well the motion decoder.
Ablations about these hyperparameters are presented in
\if\sepappendix1{Appendix~A.}
\else{Appendix~\ref{app:sec:experiments}.}
\fi

At training time, we input the full motion sequence, i.e., a variable
number of frames for each training sample. At inference time,
we can specify the desired duration $F$ (see supplementary video~\cite{projectpage_temos});
however, %
we provide quantitative metrics with known ground-truth motion duration.

\section{Experiments}
\label{sec:experiments}

We first present the data and performance measures
used in our experiments (Section~\ref{subsec:datasets}).
Next, we compare to previous work (Section~\ref{subsec:sota})
and present an ablation study (Section~\ref{subsec:ablations}).
Then, we demonstrate our results with the SMPL model (Section~\ref{subsec:smpl}).
Finally, we %
discuss limitations (Section~\ref{subsec:limitations}).

\subsection{Data and evaluation metrics}
\label{subsec:datasets}

\noindent\textbf{KIT Motion-Language~\cite{Plappert2016_KIT_ML} dataset (KIT)}
provides raw motion capture (MoCap) data, as well as processed data using the Master Motor Map (MMM) framework~\cite{mmm_terlemez}.
The motions comprise a collection of subsets of the KIT Whole-Body Human Motion Database~\cite{kit_wholebody_mandery} and of the CMU Graphics Lab Motion Capture Database~\cite{cmuWEB}.
The dataset consists of 3911 motion sequences
with 6353 sequence-level description annotations, with 9.5 words per
description on average. We use the same splits as in Language2Pose~\cite{Ahuja2019Language2PoseNL} by extracting 1784 training, 566 validation and 587 test motions
(some motions do not have corresponding descriptions).
As the model from Ghosh et al.~\cite{Ghosh_2021_ICCV} produce only 520 sequences in the test set (instead of 587), for a fair comparison we evaluate all methods with this subset, which we will refer to as the test set.
If the same motion sequence corresponds to multiple descriptions, we randomly choose
one of these descriptions at each training iteration, while we evaluate the method on the first description.
Recent state-of-the-art methods on text-conditioned motion synthesis
employ this dataset, by first converting the MMM axis-angle data into
21 $xyz$ coordinates and downsampling the sequences from 100 Hz
to 12.5 Hz. We do the same procedure, and follow the training and test splits explained above to compare methods.
Additionally, we find correspondences from the KIT sequences
to the AMASS MoCap collection~\cite{AMASS:ICCV:2019} to obtain the motions in SMPL
body format. We note that this procedure resulted in
a subset of 2888 annotated motion sequences, as some sequences have not been processed in AMASS.
We refer to this data as KIT$_{\text{SMPL}}$.

\noindent\textbf{Evaluation metrics.}
We follow the performance measures employed in
Language2Pose~\cite{Ahuja2019Language2PoseNL} and
Ghosh~et~al.~\cite{Ghosh_2021_ICCV} for quantitative evaluations.
In particular, we report Average Positional Error (APE) and Average Variance Error (AVE) metrics.
However, we note that the results in \cite{Ghosh_2021_ICCV}
do not match the ones in \cite{Ahuja2019Language2PoseNL} due to lack
of evaluation code from \cite{Ahuja2019Language2PoseNL}.
We identified minor issues with the evaluation code of \cite{Ghosh_2021_ICCV}
(more details in \if\sepappendix1{Appendix~C);}
\else{Appendix~\ref{app:sec:evaluation_metrics});}
\fi
therefore, we reimplement our own evaluation.
Moreover, we introduce several modifications (which we believe make the metrics
more interpretable):
in contrast to \cite{Ghosh_2021_ICCV,Ahuja2019Language2PoseNL},
we compute the root joint metric by using the joint coordinates only
(and not on velocities for $x$ and $y$ axes) and all the metrics are computed without standardizing the data (i.e., mean subtraction and division by standard deviation).
Our motivation for this is to remain in the coordinate space
since the metrics are \textit{positional}.
Note that the KIT data in the MMM format is canonicalized to the same body shape.
We perform most of our experiments with this data format to remain comparable
to the state of the art. We report results with the SMPL body format
separately since the skeletons are not perfectly compatible (see \if\sepappendix1{Appendix~A.5).}
\else{Appendix~\ref{app:sec:quantitative_smpl}).}
\fi
Finally, we convert our low-fps generations (at 12 Hz) to the original frame-rate of KIT (100 Hz)
via linear interpolation on coordinates and report the error comparing to this original ground truth.
We display the error in meters.

As discussed in Section~\ref{sec:intro},
the evaluation is suboptimal because it assumes one ground truth motion per text;
however, our focus is to generate multiple different motions.
The KIT test set is insufficient to design distribution-based
metrics such as FID, since there are not enough motions
for the same text (see \if\sepappendix1{Appendix~E.}
\else{Appendix~\ref{app:sec:statistics}}
\fi for statistics).
We therefore report the performance of
generating a single sample, as well as generating
multiple and evaluating the closest sample to the ground truth.
We rely on additional perceptual studies to assess
the correctness of multiple generations, which is described in \if\sepappendix1{Appendix~C.}
\else{Appendix~\ref{app:sec:evaluation_metrics}.}
\fi

\subsection{Comparison to the state of the art}
\label{subsec:sota}
\noindent\textbf{Quantitative.} We compare with the state-of-the-art text-conditioned
motion generation methods \cite{lin2018,Ahuja2019Language2PoseNL,Ghosh_2021_ICCV} on the test set of the KIT dataset (as defined in \ref{subsec:datasets}).
To obtain motions for these three methods,
we use their publicly available codes
(note that all three give the ground truth initial frame as
input to their generations).
We summarize the main results in Table~\ref{tab:sota}.
Our \methodshort{} approach substantially outperforms
on all metrics, except APE on local joints.
As pointed by \cite{Ahuja2019Language2PoseNL,Ghosh_2021_ICCV},
the most difficult metric that better differentiates improvements on this dataset
is the APE on the root joint, and we obtain significant improvements
on this metric. Moreover, we sample a random latent vector
for reporting the results for \methodshort{}; however, as we will show next in Section~\ref{subsec:ablations},
if we sample more, we are more likely to find the motion closer to the
ground truth.

\noindent\textbf{Qualitative.} We further provide qualitative comparisons in Figure~\ref{fig:sota}
with the state of the art.
We show sample generations for Lin~et~al.~\cite{lin2018},
JL2P~\cite{Ahuja2019Language2PoseNL}, and
Ghosh~et~al.~\cite{Ghosh_2021_ICCV}.
The motions from our \methodshort{} model
reflect the semantic content of the input text
better than the others across a variety of samples.
Furthermore,
we observe that
while \cite{lin2018} generates overly smooth motions,
JL2P has lots of foot sliding. \cite{Ghosh_2021_ICCV},
on the other hand, synthesizes unrealistic motions
due to exaggerated foot contacts (and even extremely elongated limbs such as in 3rd column, 3rd row of Figure~\ref{fig:sota}).
Our generations are the most realistic among all.
Further visualizations are provided in the supplementary video~\cite{projectpage_temos}.

\noindent\textbf{Perceptual study.}
These conclusions are further justified by two human perceptual studies that evaluate which methods are preferred in terms of semantics (correspondence to the text) or in terms of realism.
For the first study, we displayed a pair of motions (with a randomly swapped order
in each display) and a description of the motion, %
and asked Amazon
Mechanical Turk (AMT) workers the question:
``Which motion corresponds better to the textual description?''.
We collected answers for 100 randomly sampled test descriptions,
showing each description to multiple workers. 
For the second study, we asked another set of AMT workers the question: ``Which motion is more realistic?'' without showing the description.
We give more details
on our perceptual studies in \if\sepappendix1{Appendix~C.}
\else{Appendix~\ref{app:sec:evaluation_metrics}.}
\fi

\begin{table*}[t]
    \centering
    \caption{\textbf{State-of-the-art comparison:}
        We compare our method with recent works~\cite{lin2018,Ahuja2019Language2PoseNL,Ghosh_2021_ICCV}, on the KIT Motion-Language dataset~\cite{Plappert2016_KIT_ML} and obtain significant improvements
        on most metrics (values in meters) even if we are sampling a random motion per text conditioning
        for our model.
    }
    \setlength{\tabcolsep}{4pt}
    \resizebox{0.99\linewidth}{!}{
    \begin{tabular}{l|cccc|cccc}
        \toprule
         \multirow{2}{*}{\textbf{Methods}} & \multicolumn{4}{c|}{Average Positional Error $\downarrow$} & \multicolumn{4}{c}{Average Variance Error $\downarrow$} \\
         & \small{root joint} & \small{global traj.} & \small{mean local} & \small{mean global} &
         \small{root joint} & \small{global traj.} & \small{mean local} & \small{mean global}  \\
    \midrule
\textbf{Lin et. al.~\cite{lin2018}} & 1.966 & 1.956 & 0.105 & 1.969 & 0.790 & 0.789 & 0.007 & 0.791 \\
\textbf{JL2P~\cite{Ahuja2019Language2PoseNL}} & 1.622 & 1.616 & \textbf{0.097} & 1.630 & 0.669 & 0.669 & 0.006 & 0.672 \\
\textbf{Ghosh et al.~\cite{Ghosh_2021_ICCV}} & 1.291 & 1.242 & 0.206 & 1.294 & 0.564 & 0.548 & 0.024 & 0.563 \\
\textbf{TEMOS (ours)} & \textbf{0.963} & \textbf{0.955} & 0.104 & \textbf{0.976} & \textbf{0.445} & \textbf{0.445} & \textbf{0.005} & \textbf{0.448} \\
     \bottomrule
    \end{tabular}
    }
    \label{tab:sota}
\end{table*}

\begin{figure}[t]
    \centering
    \includegraphics[width=0.99\linewidth]{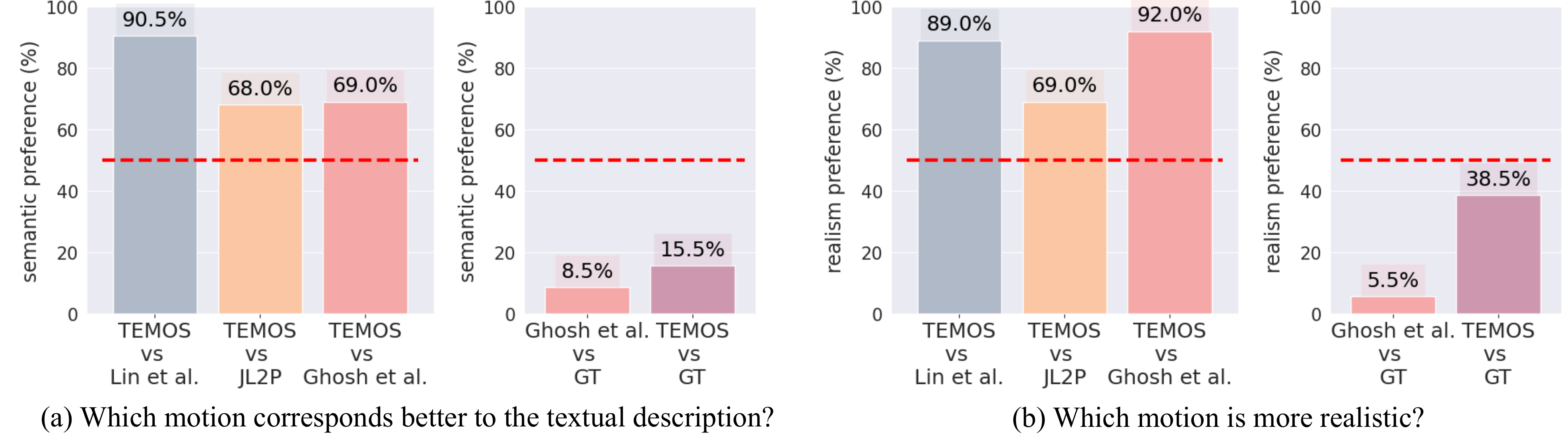}
    \caption{\textbf{Perceptual study}: (a) We ask users
    which motion corresponds better to the input text
    between two displayed samples generated
    from model A vs model B. (b) We ask other users which motion is more realistic without showing the textual description.
    We report
    the percentage for which the users
    show a preference for A.
    The red dashed line denotes
    the 50\% level (equal preference).
    On the left of both studies,
    our generations from TEMOS were rated
    better than the previous work of
    Lin~et~al.~\cite{lin2018},
    JL2P~\cite{Ahuja2019Language2PoseNL}, and
    Ghosh~et~al.~\cite{Ghosh_2021_ICCV}.
    On the right of both studies, we compare against the ground truth (GT)
    and see that our motions are rated
    as better than the GT 15.5\% and 38.5\% of the time, whereas
    Ghosh~et~al.~\cite{Ghosh_2021_ICCV} are at 8.5\% and 5.5\%.}
    \label{fig:humanstudy}
\end{figure}

The resulting ranking between our method and
each of the state-of-the-art methods \cite{lin2018,Ahuja2019Language2PoseNL,Ghosh_2021_ICCV} is reported
in Figure~\ref{fig:humanstudy}.
We see that humans perceive our motions as
better matching the descriptions
compared to all three state-of-the-art methods,
especially significantly outperforming Lin~et~al.~\cite{lin2018}
(users preferred \methodshort{} over \cite{lin2018} 90.5\% of the time).
For the more competitive and more recent Ghosh~et~al.~\cite{Ghosh_2021_ICCV}
method, we ask users to compare their generations against the ground truth.
We do the same for our generations and see that
users preferred our motions over the ground truth 15.5\% of the time
where the ones from Ghosh~et~al.~\cite{Ghosh_2021_ICCV}
are preferred only 8.5\% of the time.
Our generations are also clearly preferred in terms of realism over the three methods.
Our motions are realistic enough that they are preferred to real motion capture data 38.5\% of the time, as compared to 5.5\% of the time for Ghosh~et~al.~\cite{Ghosh_2021_ICCV}.

\begin{figure}[t]
    \centering
    \includegraphics[width=.99\textwidth]{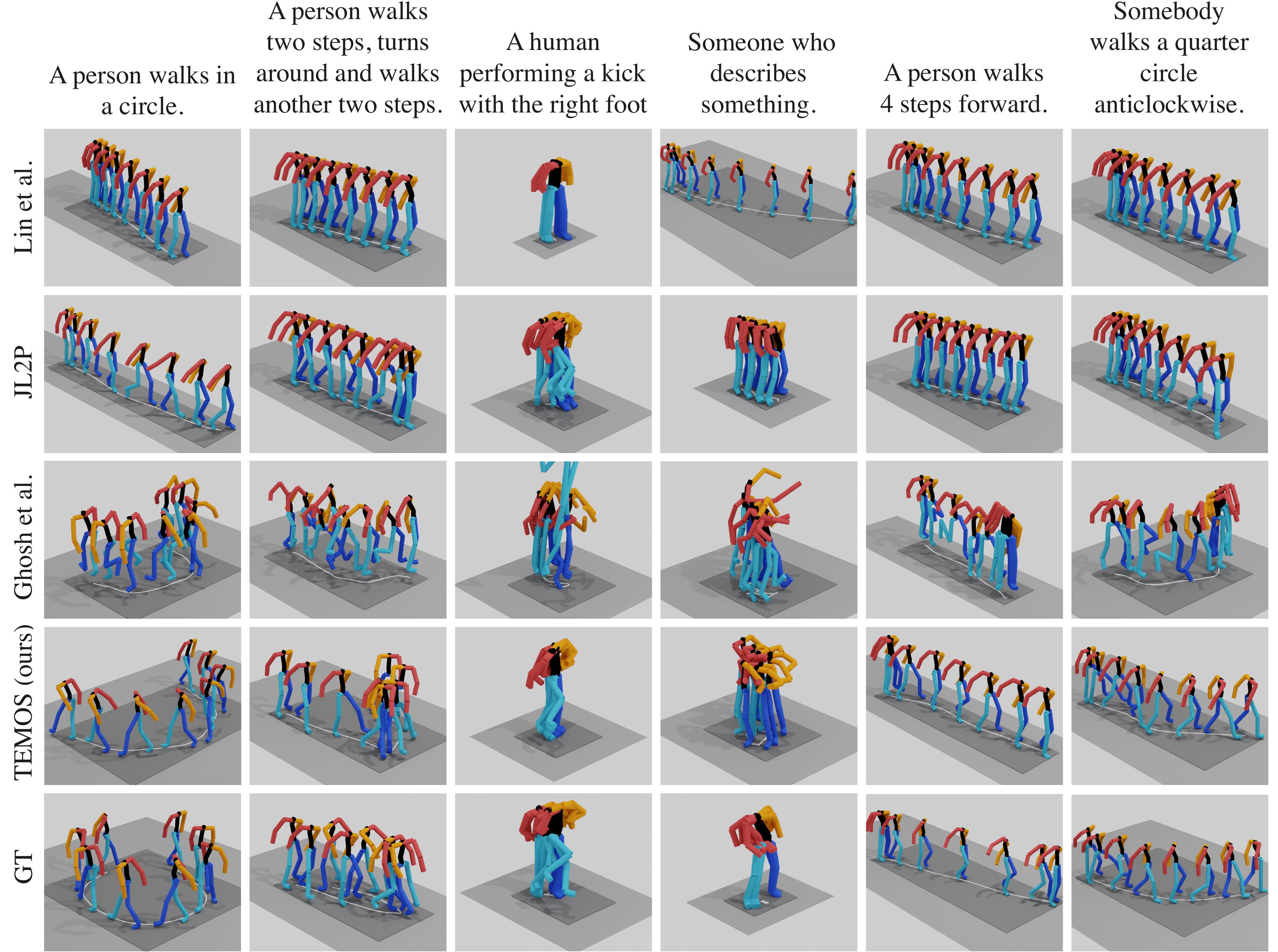}
    \caption{\textbf{Qualitative comparison to the state of the art}:
        We qualitatively compare the generations from our \methodshort{}~model with the recent state-of-the-art methods and the ground truth (GT). We present different
        textual queries in columns, and different methods in rows.
        Overall, our generations better match semantically to
        the textual descriptions.
        We further overcome
        several limitations with the prior work, such as over-smooth
        motions in Lin~et~al.~\cite{lin2018}, foot sliding in J2LP~\cite{Ahuja2019Language2PoseNL},
        and exaggerated foot contacts in Ghosh~et~al.~\cite{Ghosh_2021_ICCV},
        which can better be viewed in our supplementary video~\cite{projectpage_temos}.
    }
    \label{fig:sota}
\end{figure}

\begin{table*}[t]
    \centering
    \caption{\textbf{Variational vs deterministic models:}
    We first provide the performance of the deterministic version of our model.
    We then report results with several settings using our variational model: (i) generating a single
    motion per text to compare against the ground truth (either randomly or using a zero-vector representing
    the mean of the Gaussian latent space), and (ii) generating 10 motions per text,
    each compared against the ground truth separately (either averaging the metrics or taking the motion
    with the best metric). As expected, TEMOS is able to produce multiple hypotheses where
    the best candidates improve the metrics.
    }
    \setlength{\tabcolsep}{4pt}
    \resizebox{0.99\linewidth}{!}{
    \begin{tabular}{ll|cccc|cccc}
        \toprule
         \multirow{2}{*}{\textbf{Model}} & \multirow{2}{*}{\textbf{Sampling}} & \multicolumn{4}{c|}{Average Positional Error $\downarrow$} & \multicolumn{4}{c}{Average Variance Error $\downarrow$} \\
          & & \small{root joint} & \small{global traj.} & \small{mean local} & \small{mean global} &
         \small{root joint} & \small{global traj.} & \small{mean local} & \small{mean global}  \\
    \midrule

Deterministic & n/a & 1.175 & 1.165 & 0.106 & 1.188 & 0.514 & 0.513 & 0.005 & 0.516 \\
\midrule
Variational & 1 sample, $z=\vec{0}$ & 1.005 & 0.997 & 0.104 & 1.020 & 0.443 & 0.442 & 0.005 & 0.446 \\
Variational & 1 random sample & 0.963 & 0.955 & 0.104 & 0.976 & 0.445 & 0.445 & 0.005 & 0.448 \\
Variational & 10 random avg & 1.001 & 0.993 & 0.104 & 1.015 & 0.451 & 0.451 & 0.005 & 0.454 \\
Variational & 10 random best & \textbf{0.784} & \textbf{0.774} & 0.104 & \textbf{0.802} & \textbf{0.392} & \textbf{0.391} & 0.005 & \textbf{0.395} \\
     \bottomrule
    \end{tabular}
    }
    \label{tab:variational}
\end{table*}

\subsection{Ablation study}
\label{subsec:ablations}
In this section, we evaluate the influence of several components of our
framework in a controlled setting.

\noindent\textbf{Variational design.}
First, we `turn off' the variational property of
our generative model and synthesize a single motion
per text. Instead of two learnable distribution tokens as in Figure~\ref{fig:model},
we use one learnable \textit{embedding} token from which we directly
obtain the latent vector using the corresponding encoder output (hence removing
sampling).
We removed all the KL losses such that the model becomes deterministic, and keep the embedding similarity loss to learn the joint latent space.
In Table~\ref{tab:variational},
we report performance metrics with this approach
and see that we already obtain competitive performance
with the deterministic version of our model,
demonstrating the improvements from our temporal
sequence modeling approach compared to previous works.

As noted earlier, our variational model
can produce multiple generations for the same text,
and a single random sample may not necessarily
match the ground truth.
In Table~\ref{tab:variational}, we report results
for one generation from a random $z$ noise vector,
or generating from the zero-vector that represents
the mean of the latent space ($z=\vec{0}$);
both perform similarly.
To assess the performance with multiple generations,
we randomly sample 10
latent vectors per text, and provide two evaluations.
First, we compare each of the 10 generations to the
single ground truth, and average over
all generations (10 random avg).
Second, we record the performance of the
motion that best matches to the ground truth
out of the 10 generations (10 random best).
As expected, Table~\ref{tab:variational}
shows improvements with the latter (see \if\sepappendix1{Appendix~A.5}
\else{Appendix~\ref{app:subsec:vae_sampling}}
\fi
for more values for the number of latent vectors).

\noindent\textbf{Architectural and loss components.}
Next, we investigate which component is most responsible
for the performance improvement over the state of the art,
since even the deterministic variant of our model outperforms
previous works. Table~\ref{tab:architecture}
reports the performance by removing one component at each row.
The APE root joint performance drops from 0.96 to
i) 1.44 using GRUs instead of Transformers;
ii) 1.18 without the motion encoder (using only one KL loss);
iii) 1.09 without the cross-modal embedding loss;
iv) 1.05 without the Gaussian priors;
v) 0.99 without the cross-modal KL losses.
Note that the cross-modal framework originates from
JL2P~\cite{Ahuja2019Language2PoseNL}. While we observe
slight improvement with each of the cross-modal terms,
we notice that the model performance is already satisfactory
even without the motion encoder.
We therefore conclude that the main improvement stems from the improved non-autoregressive
Transformer architecture, and removing each of the other components (4 KL
loss terms, motion encoder, embedding similarity) also slightly degrades performance.

\begin{table*}[t]
    \centering
    \caption{\textbf{Architectural and loss study:}
        We conclude that the most critical component is the Transformer
        architecture, as opposed to a recurrent one (i.e., GRU). While
        the additional losses are helpful, they bring relatively minor improvements.
    }
    \setlength{\tabcolsep}{4pt}
    \resizebox{0.99\linewidth}{!}{
        \begin{tabular}{lll|cccc|cccc}
            \toprule
             & & & \multicolumn{4}{c|}{Average Positional Error $\downarrow$} & \multicolumn{4}{c}{Average Variance Error $\downarrow$} \\
            & & & root & glob. & mean & mean & root & glob. & mean & mean \\
            Arch. & $\mathcal{L}_{KL}$ & $\mathcal{L}_{E}$ & joint & traj. & loc. & glob. & joint & traj. & loc. & glob.  \\
            \midrule
GRU & $KL(\phi^T, \phi^M) + KL(\phi^M, \phi^T) + KL(\phi^T, \psi) + KL(\phi^M, \psi)$ & \cmark & 1.443 & 1.433 & 0.105 & 1.451 & 0.600 & 0.599 & 0.007 & 0.601 \\
\midrule
Transf. & $KL(\phi^T, \psi)$ \textbf{w/out $\mathscr{M}_{enc}$} & \xmark & 1.178 & 1.168 & 0.106 & 1.189 & 0.506 & 0.505 & 0.006 & 0.508 \\
Transf. & $KL(\phi^T, \phi^M) + KL(\phi^M, \phi^T) + KL(\phi^T, \psi) + KL(\phi^M, \psi)$  & \xmark & 1.091 & 1.083 & 0.107 & 1.104 & 0.449 & 0.448 & 0.005 & 0.451 \\
Transf. & $KL(\phi^T, \psi) + KL(\phi^M, \psi)$ \textbf{w/out cross-modal KL losses} & \xmark & 1.080 & 1.071 & 0.107 & 1.095 & 0.453 & 0.452 & 0.005 & 0.456 \\
Transf. & $KL(\phi^T, \psi) + KL(\phi^M, \psi)$ \textbf{w/out cross-modal KL losses} & \cmark & 0.993 & 0.983 & 0.105 & 1.006 & 0.461 & 0.460 & 0.005 & 0.463 \\
Transf. & $KL(\phi^T, \phi^M) + KL(\phi^M, \phi^T)$ \textbf{w/out Gaussian priors} & \cmark & 1.049 & 1.039 & 0.108 & 1.065 & 0.472 & 0.471 & 0.005 & 0.475 \\
\midrule
Transf. & $KL(\phi^T, \phi^M) + KL(\phi^M, \phi^T) + KL(\phi^T, \psi) + KL(\phi^M, \psi)$ & \cmark & \textbf{0.963} & \textbf{0.955} & \textbf{0.104} & \textbf{0.976} & \textbf{0.445} & \textbf{0.445} & 0.005 & \textbf{0.448} \\
            \bottomrule
        \end{tabular}
    }
    \label{tab:architecture}
\end{table*}

\noindent\textbf{Language model finetuning.}
As explained in Section~\ref{subsec:envae},
we do not update the language model parameters
during training, which are from the pretrained DistilBERT~\cite{distilbert_sanh}.
We measure the performance with and without finetuning
in Table~\ref{tab:ablations}
and conclude that freezing performs better while being
more efficient. We note that we already introduce
additional layers through our text encoder (see Figure~\ref{fig:model}),
which may be sufficient to adapt the embeddings to our specific motion description domain.
We provide an additional experiment with larger language models in \if\sepappendix1{Appendix~A.4.}
\else{Appendix~\ref{app:subsec:lm}.}
\fi

\begin{table*}[t]
    \centering
    \caption{\textbf{Language model finetuning:}
        We experiment with finetuning the language model (LM) parameters 
        (i.e., DistilBERT~\cite{distilbert_sanh})
        end-to-end with our motion-language cross-modal framework,
        and do not observe improvements. Here `Frozen' refers
        to not updating the LM parameters.
    }
    \setlength{\tabcolsep}{6pt}
    \resizebox{0.99\linewidth}{!}{
    \begin{tabular}{l|cccc|cccc}
        \toprule
         \multirow{2}{*}{\textbf{LM params}} & \multicolumn{4}{c|}{Average Positional Error $\downarrow$} & \multicolumn{4}{c}{Average Variance Error $\downarrow$} \\
         & \small{root joint} & \small{global traj.} & \small{mean local} & \small{mean global} &
         \small{root joint} & \small{global traj.} & \small{mean local} & \small{mean global} \\
    \midrule
Finetuned & 1.402 & 1.393 & 0.113 & 1.414 & 0.559 & 0.558 & 0.006 & 0.562 \\
Frozen & \textbf{0.963} & \textbf{0.955} & \textbf{0.104} & \textbf{0.976} & \textbf{0.445} & \textbf{0.445} & \textbf{0.005} & \textbf{0.448} \\
     \bottomrule
    \end{tabular}
    }
    \label{tab:ablations}
\end{table*}

\begin{figure}[t]
    \centering
    \includegraphics[width=0.97\textwidth]{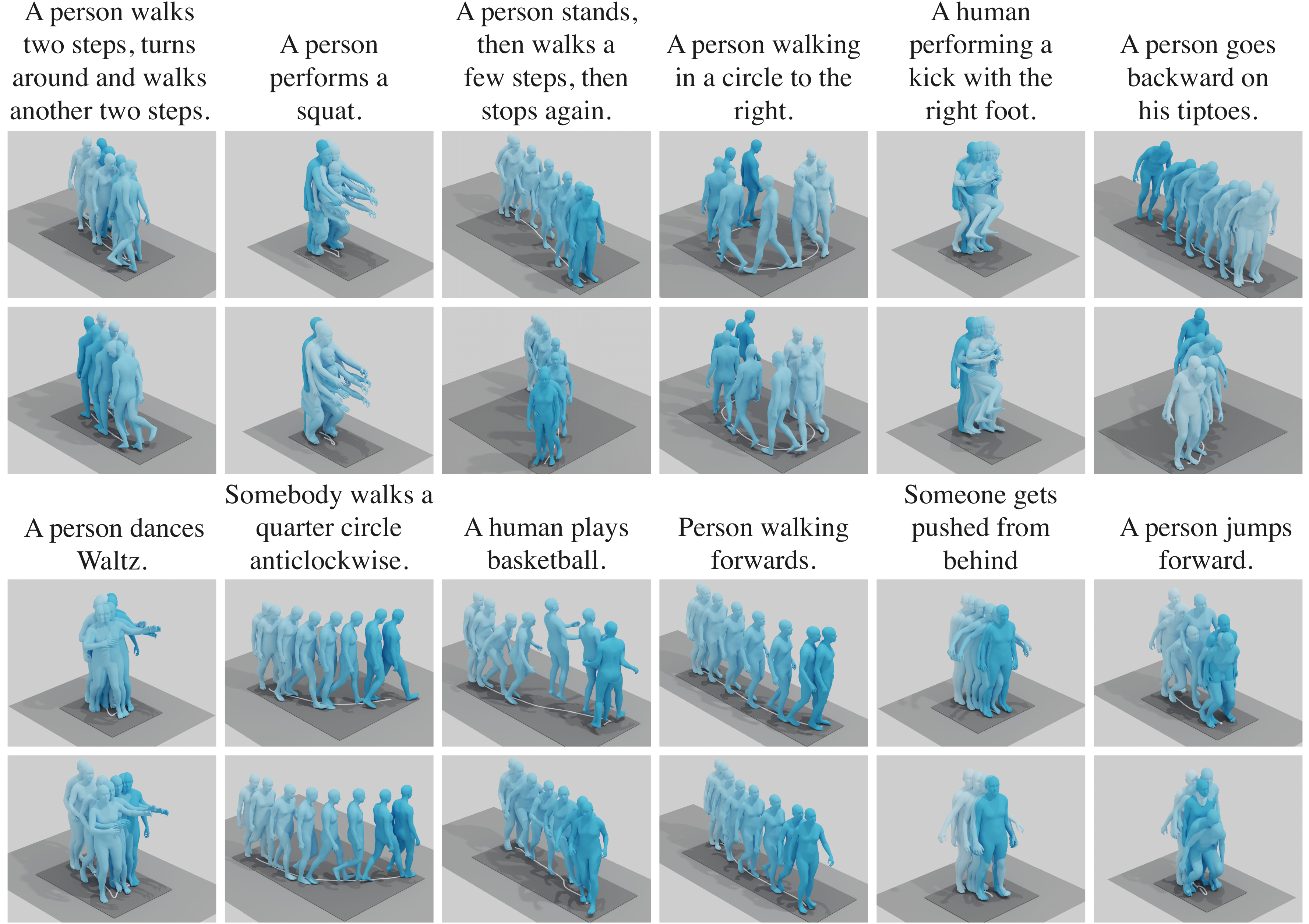}
    \caption{\textbf{Qualitative evaluation of the diversity}:
    We display two motion generations for each description.
    Our model shows certain diversity among different generations
    while respecting the textual description.
    }
    \label{fig:qualitative}
\end{figure}

\subsection{Generating skinned motions}
\label{subsec:smpl}
We evaluate the variant of our model which
uses the parametric SMPL representation to
generate full body meshes. The quantitative performance metrics
on KIT$_{\text{SMPL}}$ test set can be found in \if\sepappendix1{Appendix~A.5.}
\else{Appendix~\ref{app:sec:quantitative_smpl}.}
\fi
We provide qualitative examples
in Figure~\ref{fig:qualitative}
to illustrate the diversity of our generations for a given text.
For each text, we present 2 random samples.
Each column shows a different text input. For all the visualization
renderings in this paper, the camera is fixed and the bodies are sampled evenly across time.
Moreover, the forward direction of the first frame is always facing the same canonical direction.
Our observation is that the model can generate
multiple plausible motions corresponding to the same
text, exploring the degrees of freedom remaining
from ambiguities in the language description.
On the other hand, if the text describes a precise action,
such as `A person performs a squat'
the diversity is reduced.
The results are better seen as movies; see supplementary video~\cite{projectpage_temos},
where we also display other effects such as generating variable durations,
and interpolating in the latent space.%

\subsection{Limitations}
\label{subsec:limitations}

Our model has several limitations.
Firstly, the vocabulary of the KIT data is relatively
small with 1263 unique words
compared to the full 
open-vocabulary setting of
natural language, and are dominated by locomotive motions.
We therefore expect our model
to suffer from out-of-distribution descriptions.
Moreover, we do not have a principled way of
measuring the diversity of our models since
the training does not include multiple motions
for the exact same text.
Secondly, we notice that if the input text %
contains typos (e.g., `wals' instead of `walks'),
\methodshort{} %
might drastically fail,
suggesting that a preprocessing step to correct them beforehand 
might be needed.
Finally, our method 
cannot scale up to very long motions %
(such as walking for several minutes) due to the quadratic memory cost. %

\section{Conclusion}

In this work, we introduced a variational approach
to generate diverse 3D human motions given
textual descriptions in the form of natural language.
In contrast to previous methods, our approach
considers the intrinsically ambiguous nature
of language and generates multiple plausible
motions respecting the textual description,
rather than deterministically producing only one.
We obtain state-of-the-art results on the
widely used KIT Motion-Language benchmark,
outperforming prior work by a large margin
both in quantitative experiments and perceptual studies.
Our improvements are mainly from the
architecture design of incorporating sequence modeling
via Transformers. Furthermore,
we employ full body meshes instead of only skeletons.
Future work should focus on explicit
modeling of contacts and integrating physics knowledge.
Another interesting direction is to explore
duration estimation for the generated motions.
While we do not expect any immediate negative
societal impact from our work, we note that
with the potential advancement of fake
visual data generation, a risk
may arise from the integration of our model
in the applications that animate existing people without their consent,
raising privacy concerns.
\bigskip

{ \small
\noindent\textbf{Acknowledgements.}
This work was granted access to
the HPC resources of IDRIS under the allocation 2021-AD011012129R1 made by GENCI.
GV 
acknowledges the ANR project CorVis ANR-21-CE23-0003-01, and research gifts from Google and Adobe.
The authors would like to thank Monika Wysoczanska, Georgy Ponimatkin, Romain Loiseau, Lucas Ventura and Margaux Lasselin for their feedbacks, Tsvetelina Alexiadis and Taylor McConnell for helping with the perceptual study, and Anindita Ghosh for helping with the evaluation details on the KIT dataset.
\par
}
{ \small
\noindent\textbf{Disclosure:} \url{https://files.is.tue.mpg.de/black/CoI_ECCV_2022.txt}
\par
}

\bibliographystyle{splncs04}
\bibliography{references}

\newpage
{\noindent \large \bf {APPENDIX}}\\
\renewcommand{\thefigure}{A.\arabic{figure}} %
\setcounter{figure}{0} 
\renewcommand{\thetable}{A.\arabic{table}}
\setcounter{table}{0} 

\appendix

This appendix provides additional experiments (Section~\ref{app:sec:experiments}),
description of the motion representations (Section~\ref{app:sec:representation}),
details on evaluation metrics (Section~\ref{app:sec:evaluation_metrics}),
implementation details (Section~\ref{app:sec:implementation}),
and dataset statistics (Section~\ref{app:sec:statistics}).

\subsubsection{Video.} Additionally, we provide a supplemental video, available on our website~\cite{projectpage_temos}, which we encourage the reader to watch since motion is critical in our results, and this is hard to convey in a static document.
In the video, we illustrate: (i) comparison with previous work, (ii) training with skeleton versus SMPL data, (iii) diversity of our model, (iv) generation of variable size sequences,
(v) interpolation between two texts in our latent space, and (vi) failure cases.

\subsubsection{Code.} We also share our code base
on the project page~\cite{projectpage_temos}, which reproduces our training and evaluation metrics. Explanations on how to configure the data and launch the training can be found in the file \texttt{README.md}.

\section{Additional experiments}
\label{app:sec:experiments}
We conduct several experiments to explore the sensitivity of our model
to certain hyperparameters. 
Our final set of hyperparameters was chosen using the validation set (starting from a set of hyperparameters similar to ACTOR~\cite{petrovich21actor}). Note that these hyperparameters did not always appear optimal on the test set. The following ablations provide a sense of robustness to different parameters.
In particular,
we show the effects of 
the batch size (Section~\ref{app:subsec:batchsize}), $\{\lambda_{\text{KL}}, \lambda_{\text{E}}\}$
loss weighting parameters (Section~\ref{app:subsec:kl})
and the architecture parameters of Transformers (Section~\ref{app:subsec:trans}). All other parameters, such as the learning rate ($10^{-4}$), the optimizer (AdamW) and the number of epochs ($1000$) are fixed. For all these experiments we use the skeleton-based
MMM representation.

We also experiment with various pretrained language models (Section~\ref{app:subsec:lm}).

Note that the evaluation is based on a single random sample. All results can be improved by taking the best sample closest to the ground truth out of multiple generations (as next explained in Section~\ref{app:subsec:vae_sampling}). We also experiment with taking the sample farthest from the ground truth.

Finally, in Section~\ref{app:sec:quantitative_smpl}, we report quantitative results for our
model trained with SMPL rotations.

\subsection{Batch size}
\label{app:subsec:batchsize}
\begin{table*}[b]
    \centering
    \caption{\textbf{Batch size:}
        We see that the performance is the best for either a small batch size (=8) or a bigger batch size (=32). We were unable to use a 
        higher batch size due to the GPU memory limit. %
    }
    \setlength{\tabcolsep}{4pt}
    \resizebox{0.99\linewidth}{!}{
    \begin{tabular}{l|cccc|cccc}
        \toprule
         \multirow{2}{*}{\textbf{Batch size}} & \multicolumn{4}{c|}{Average Positional Error $\downarrow$} & \multicolumn{4}{c}{Average Variance Error $\downarrow$} \\
         & \small{root joint} & \small{global traj.} & \small{mean local} & \small{mean global} &
         \small{root joint} & \small{global traj.} & \small{mean local} & \small{mean global} \\
    \midrule
bs = 8 & \textbf{0.950} & \textbf{0.941} & 0.105 & \textbf{0.965} & 0.449 & 0.448 & 0.005 & 0.451 \\
bs = 16 & 1.115 & 1.106 & 0.105 & 1.128 & 0.513 & 0.512 & 0.005 & 0.515 \\
bs = 24 & 1.260 & 1.250 & 0.106 & 1.273 & 0.542 & 0.542 & 0.005 & 0.545 \\
bs = 32 & 0.963 & 0.955 & \textbf{0.104} & 0.976 & \textbf{0.445} & \textbf{0.445} & 0.005 & \textbf{0.448} \\

     \bottomrule
    \end{tabular}
    }
    \label{tab:bs_ablations}
\end{table*}

\begin{table*}[t]
    \centering
    \caption{\textbf{Weight of the KL losses ($\lambda_{\text{KL}}$) and the embedding loss ($\lambda_{\text{E}}$):} The results are influenced more by changes in $\lambda_{\text{E}}$ than in $\lambda_{\text{KL}}$, but otherwise if the values are not too low, the performances are similar. Note that the control row
    $\lambda_{\text{KL}}=10^{-5}, \lambda_{\text{E}}=10^{-5}$ is repeated in each block.
    }
    \setlength{\tabcolsep}{4pt}
    \resizebox{0.99\linewidth}{!}{
    \begin{tabular}{l|cccc|cccc}
        \toprule
         \multirow{2}{*}{\textbf{Losses weight}} & \multicolumn{4}{c|}{Average Positional Error $\downarrow$} & \multicolumn{4}{c}{Average Variance Error $\downarrow$} \\
         & \small{root joint} & \small{global traj.} & \small{mean local} & \small{mean global} &
         \small{root joint} & \small{global traj.} & \small{mean local} & \small{mean global} \\
    \midrule
$\lambda_{\text{KL}}=\lambda_{\text{E}}=10^{-3}$ & 1.219 & 1.210 & 0.111 & 1.230 & 0.555 & 0.554 & 0.006 & 0.556 \\
$\lambda_{\text{KL}}=\lambda_{\text{E}}=10^{-4}$ & 1.110 & 1.101 & 0.106 & 1.122 & 0.476 & 0.475 & 0.005 & 0.479 \\
$\lambda_{\text{KL}}=\lambda_{\text{E}}=10^{-5}$ & \textbf{0.963} & \textbf{0.955} & \textbf{0.104} & \textbf{0.976} & \textbf{0.445} & \textbf{0.445} & 0.005 & \textbf{0.448} \\
$\lambda_{\text{KL}}=\lambda_{\text{E}}=10^{-6}$ & 1.242 & 1.233 & 0.105 & 1.254 & 0.586 & 0.585 & 0.005 & 0.589 \\
$\lambda_{\text{KL}}=\lambda_{\text{E}}=10^{-7}$ & 1.034 & 1.025 & 0.108 & 1.049 & 0.488 & 0.487 & 0.005 & 0.491 \\
$\lambda_{\text{KL}}=\lambda_{\text{E}}=10^{-8}$ & 1.085 & 1.075 & 0.107 & 1.099 & 0.490 & 0.489 & 0.005 & 0.493 \\
\midrule
$\lambda_{\text{KL}}=10^{-5}, \lambda_{\text{E}}=10^{-3}$ & 1.293 & 1.284 & 0.107 & 1.305 & 0.631 & 0.631 & 0.005 & 0.635 \\
$\lambda_{\text{KL}}=10^{-5}, \lambda_{\text{E}}=10^{-4}$ & 1.039 & 1.029 & 0.104 & 1.052 & 0.449 & 0.448 & 0.005 & 0.452 \\
$\lambda_{\text{KL}}=10^{-5}, \lambda_{\text{E}}=10^{-5}$ & \textbf{0.963} & \textbf{0.955} & \textbf{0.104} & \textbf{0.976} & \textbf{0.445} & \textbf{0.445} & 0.005 & \textbf{0.448} \\
$\lambda_{\text{KL}}=10^{-5}, \lambda_{\text{E}}=10^{-6}$ & 1.082 & 1.072 & 0.107 & 1.095 & 0.456 & 0.455 & 0.005 & 0.459 \\
$\lambda_{\text{KL}}=10^{-5}, \lambda_{\text{E}}=10^{-7}$ & 1.018 & 1.008 & 0.106 & 1.031 & 0.464 & 0.463 & 0.005 & 0.467 \\
$\lambda_{\text{KL}}=10^{-5}, \lambda_{\text{E}}=10^{-8}$ & 1.076 & 1.067 & 0.105 & 1.089 & 0.477 & 0.476 & 0.005 & 0.480 \\
\midrule
$\lambda_{\text{KL}}=10^{-3}, \lambda_{\text{E}}=10^{-5}$ & 1.145 & 1.135 & 0.111 & 1.157 & 0.507 & 0.506 & 0.006 & 0.509 \\
$\lambda_{\text{KL}}=10^{-4}, \lambda_{\text{E}}=10^{-5}$ & 1.070 & 1.061 & 0.106 & 1.083 & 0.471 & 0.470 & 0.005 & 0.474 \\
$\lambda_{\text{KL}}=10^{-5}, \lambda_{\text{E}}=10^{-5}$ & \textbf{0.963} & \textbf{0.955} & \textbf{0.104} & \textbf{0.976} & \textbf{0.445} & \textbf{0.445} & 0.005 & \textbf{0.448} \\
$\lambda_{\text{KL}}=10^{-6}, \lambda_{\text{E}}=10^{-5}$ & 0.971 & 0.962 & 0.105 & 0.986 & 0.455 & 0.454 & 0.005 & 0.457 \\
$\lambda_{\text{KL}}=10^{-7}, \lambda_{\text{E}}=10^{-5}$ & 1.140 & 1.132 & 0.106 & 1.154 & 0.513 & 0.512 & 0.005 & 0.517 \\
$\lambda_{\text{KL}}=10^{-8}, \lambda_{\text{E}}=10^{-5}$ & 1.025 & 1.015 & 0.107 & 1.039 & 0.461 & 0.460 & 0.005 & 0.464 \\
     \bottomrule
    \end{tabular}
    }
    \label{tab:kl_ablations}
\end{table*}

\begin{table*}[t]
    \centering
    \caption{\textbf{Number of layers and heads in all Transformers:}
        While our results are slightly better for larger models,
        we observe that the performance is not very sensitive to changes in the number of
        layers and heads in Transformers.
    }
    \setlength{\tabcolsep}{4pt}
    \resizebox{0.99\linewidth}{!}{
    \begin{tabular}{l|cccc|cccc}
        \toprule
         \multirow{2}{*}{\textbf{Transformers parameters}} & \multicolumn{4}{c|}{Average Positional Error $\downarrow$} & \multicolumn{4}{c}{Average Variance Error $\downarrow$} \\
         & \small{root joint} & \small{global traj.} & \small{mean local} & \small{mean global} &
         \small{root joint} & \small{global traj.} & \small{mean local} & \small{mean global} \\
    \midrule
nheads = nlayers = 2 & 1.194 & 1.185 & 0.107 & 1.205 & 0.546 & 0.545 & 0.006 & 0.547 \\
nheads = nlayers = 4 & 1.189 & 1.181 & 0.104 & 1.201 & 0.500 & 0.499 & 0.005 & 0.502 \\
nheads = nlayers = 6 & \textbf{0.963} & \textbf{0.955} & 0.104 & \textbf{0.976} & \textbf{0.445} & \textbf{0.445} & 0.005 & \textbf{0.448} \\
     \bottomrule
    \end{tabular}
    }
    \label{tab:nlayers_tot_ablations}
\end{table*}

\begin{table*}[t]
    \centering
    \caption{\textbf{Number of layers and heads in the Transformer of the text encoder only:}
        We fix the Transformer layers and heads of the motion encoder and motion decoder to 6 (as in the other experiments), but we only change the number of layers and heads of the \textit{text encoder} (the one on top of DistilBert). The results suggest that training a light model on top of the language model still gives descent results, but adding more layers helps.
    }
    \setlength{\tabcolsep}{4pt}
    \resizebox{0.99\linewidth}{!}{
    \begin{tabular}{l|cccc|cccc}
        \toprule
         \multirow{2}{*}{\textbf{Transformers parameters}} & \multicolumn{4}{c|}{Average Positional Error $\downarrow$} & \multicolumn{4}{c}{Average Variance Error $\downarrow$} \\
         & \small{root joint} & \small{global traj.} & \small{mean local} & \small{mean global} &
         \small{root joint} & \small{global traj.} & \small{mean local} & \small{mean global} \\
    \midrule
nheads = nlayers = 1 & 1.163 & 1.151 & 0.110 & 1.175 & 0.529 & 0.528 & 0.006 & 0.531 \\
nheads = nlayers = 2 & 1.170 & 1.161 & 0.107 & 1.182 & 0.452 & 0.451 & 0.005 & 0.454 \\
nheads = nlayers = 3 & 1.094 & 1.085 & 0.105 & 1.106 & 0.474 & 0.473 & 0.005 & 0.476 \\
nheads = nlayers = 4 & \textbf{0.916} & \textbf{0.908} & \textbf{0.104} & \textbf{0.930} & \textbf{0.440} & \textbf{0.440} & 0.005 & \textbf{0.444} \\
nheads = nlayers = 6 & 0.963 & 0.955 & \textbf{0.104} & 0.976 & 0.445 & 0.445 & 0.005 & 0.448 \\
     \bottomrule
    \end{tabular}
    }
    \label{tab:nlayers_text_ablations}
\end{table*}

In Table~\ref{tab:bs_ablations}, we present results with batch sizes of 8, 16, 24 and 32. 
To handle variable-length training, we use padding and masking in the encoders and the decoder. 
To maintain reasonable memory consumption, we discard training samples that have more than 500 frames (after sub-sampling to 12.5 Hz): 
this corresponds to about 2.3\% of the training data. 
We show that we obtain the best results by setting the batch size to 8 or 32. We set it to 32 in all other experiments as it takes less time to train.

\subsection{Weight of the KL losses and the embedding loss}
\label{app:subsec:kl}
In Table~\ref{tab:kl_ablations},
we report results by varying both $\lambda_{\text{KL}}$ and $\lambda_{\text{E}}$ parameters (described in
\if\sepappendix1{Section~3.3}
\else{Section~\ref{subsec:training}}
of the main paper),
\fi from $10^{-3}$ to $10^{-8}$. We show that overall the results are similar when $\lambda_{\text{E}}$ is fixed. 
A too high value of $10^{-3}$ deteriorates the performance. We fix both of them to $10^{-5}$.

\begin{figure}
    \centering
    \includegraphics[width=.6\linewidth]{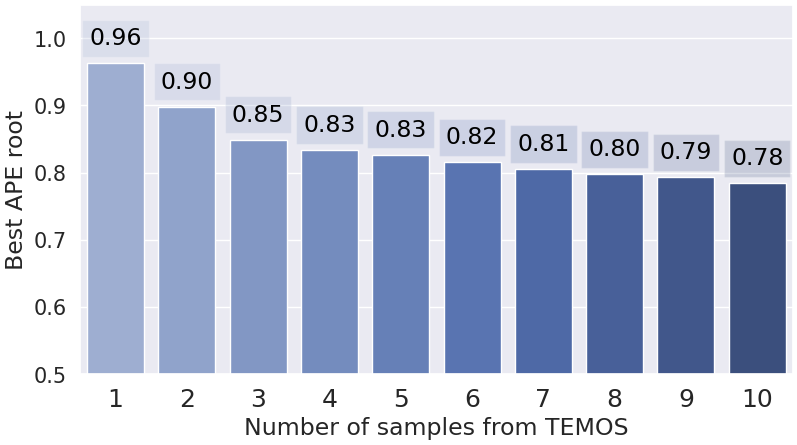}
    \caption{\textbf{Best APE root when sampling multiple generations:}
        Given a textual description, we generate multiple different motions, 
        and select the motion that matches best to the ground truth sequence. 
        We show that by sampling more generated sequences per text, 
        we can reduce the APE root metric error.}
    \label{fig:vae_best}
\end{figure}

\subsection{Number of Transformer layers}
\label{app:subsec:trans}

Here, we present two different experiments. 
The first experiment is to change globally the number of layers and the number of heads of all our Transformers.
In Table~\ref{tab:nlayers_tot_ablations},
we see that the results are optimal when they are both fixed at 6,
which is used in all other experiments.

Next, in Table~\ref{tab:nlayers_text_ablations},
we experiment with a lighter model on top of the DistilBERT text encoder,
by adding fewer layers and heads than 6. We see that 1 or 2 layers are not sufficient, but beginning with 4, the results are satisfactory. We use 6 layers in
our model.

\subsection{Pretrained language models}
\label{app:subsec:lm}
\begin{table*}[t]
    \centering
    \caption{\textbf{Language model:} We experiment with language models larger than DistilBERT and do not observe significant changes in the performance.
    }
    \setlength{\tabcolsep}{4pt}
    \resizebox{0.99\linewidth}{!}{
    \begin{tabular}{l|cccc|cccc}
        \toprule
         \multirow{2}{*}{\textbf{Language model}} & \multicolumn{4}{c|}{Average Positional Error $\downarrow$} & \multicolumn{4}{c}{Average Variance Error $\downarrow$} \\
         & \small{root joint} & \small{global traj.} & \small{mean local} & \small{mean global} &
         \small{root joint} & \small{global traj.} & \small{mean local} & \small{mean global} \\
    \midrule
DistilBERT~\cite{distilbert_sanh} & \textbf{0.963} & \textbf{0.955} & \textbf{0.104} & \textbf{0.976} & 0.445 & 0.445 & 0.005 & 0.448 \\
BERT~\cite{devlin2018bert} & 0.986 & 0.977 & 0.105 & 1.000 & \textbf{0.441} & \textbf{0.441} & 0.005 & \textbf{0.444} \\
RoBERTa~\cite{roberta2019liu} & 1.066 & 1.056 & 0.107 & 1.079 & 0.492 & 0.491 & 0.005 & 0.494 \\
     \bottomrule
    \end{tabular}
    }
    \label{tab:textmodel_ablations}
\end{table*}

We experiment with replacing DistilBERT~\cite{distilbert_sanh} with a larger pretrained language model. We compare with the original BERT~\cite{devlin2018bert} model as well as the more recent RoBERTa~\cite{roberta2019liu} model. The results are similar and suggest that DistilBERT is sufficient for this task, while having fewer parameters.

\subsection{Sampling multiple motions}
\label{app:subsec:vae_sampling}

Given a text input, instead of generating a single motion as in previous methods \cite{Ahuja2019Language2PoseNL,Ghosh_2021_ICCV,lin2018}, we can generate multiple motions.
As demonstrated in
\if\sepappendix1{Table~2}
\else{Table~\ref{tab:variational}}
\fi
of the main paper, we can improve the evaluation metrics
by picking the best out of a set of motion generations, that is closest to the ground
truth. In Figure~\ref{fig:vae_best}, we plot the reduction in APE root error
as we increase the number of generations per text from 1 to 10, and observe
a monotonic decrease as expected.

Furthermore, we measure the \textit{worst case scenario}, where we generate 10 motions per text
and record the error between the ground truth motion and the
most different generated motion out of the 10. We obtain an APE
of 1.24 (instead of 0.78 in the best case scenario, and 0.96 in the
random scenario). Note that calling this worst case may not be
accurate since the single ground truth motion does not represent
the only possible motion, i.e., our generations may correspond
well to the text without being close to the ground truth joints.

\subsection{Quantitative results with the SMPL model}
\label{app:sec:quantitative_smpl}

To evaluate quantitatively our SMPL-based model, and obtaining results comparable with the MMM framework, we extract the most similar skeleton subset from the SMPL-H joints (provided by AMASS). The correspondence can be found in Table~\ref{tab:correspondence}.

\begin{table*}[t]
    \centering
    \caption{\textbf{Correspondence} between the SMPL-H joints and the MMM framework joints.
    }
    \setlength{\tabcolsep}{4pt}
    \resizebox{0.99\linewidth}{!}{
    \begin{tabular}{l|ccccccccccc}
    \toprule
             Type & \multicolumn{11}{c}{Joints} \\
            \midrule
    MMM & root & BP & BT & BLN & BUN & LS & LE & LW & RS & RE & RW \\ 
    SMPL-H & pelvis & spine1 & spine3 & neck & head & left\_shoulder & left\_elbow & left\_wrist & right\_shoulder & right\_elbow & right\_wrist \\
    \midrule
       MMM  & LH & LK & LA & LMrot & LF & RH & RK & RA & RMrot & RF & \\
       SMPL-H & left\_hip & left\_knee & left\_ankle & left\_heel & left\_foot & right\_hip & right\_knee & right\_ankle & right\_heel & right\_foot & \\
     \bottomrule
    \end{tabular}
    }
    \label{tab:correspondence}
\end{table*}

Both in SMPL-H and MMM, the bodies are canonicalized with a standard body shape (robot-style for MMM, and average neutral body for SMPL-H). We further rescale the SMPL joints with a factor of 0.64, to match them with MMM joints. We evaluate the model with and without this rescaling. The performance metrics on KIT$_{SMPL}$ test set can be found in Table~\ref{tab:smpl_results}. Performance is comparable
to that of MMM-based training when evaluating with rescaled joints. We expect future work
to compare against the non-rescaled version when employing the SMPL model to interpret the
metrics with respect to real human sizes.

\begin{table*}[t]
    \centering
    \caption{\textbf{Our results with SMPL model:} We evaluate our model trained on the SMPL data against the ground truth of the test set of KIT$_{\text{SMPL}}$ (joints extracted from the AMASS dataset). The first row shows results after a rescaling (to get skeletons closer to MMM processed joints). The second row shows results with the same set of joints but without the final rescaling. Both results are in meters. 
    }
    \setlength{\tabcolsep}{4pt}
    \resizebox{0.99\linewidth}{!}{
        \begin{tabular}{lc|cccc|cccc}
            \toprule
             & &  \multicolumn{4}{c|}{Average Positional Error $\downarrow$} & \multicolumn{4}{c}{Average Variance Error $\downarrow$} \\
            & &  root & glob. & mean & mean & root & glob. & mean & mean \\
            Dataset & rescaled & joint & traj. & loc. & glob. & joint & traj. & loc. & glob.  \\
            \midrule
            
KIT$_{\text{SMPL}}$ & \cmark & 0.698 & 0.689 & 0.091 & 0.712 & 0.157 & 0.157 & 0.004 & 0.161 \\
KIT$_{\text{SMPL}}$ & \xmark & 1.097 & 1.077 & 0.169 & 1.118 & 0.384 & 0.383 & 0.009 & 0.393 \\
\bottomrule
        \end{tabular}
    }
    \label{tab:smpl_results}
\end{table*}

\begin{figure}[h]
    \centering
    \includegraphics[width=.20\linewidth]{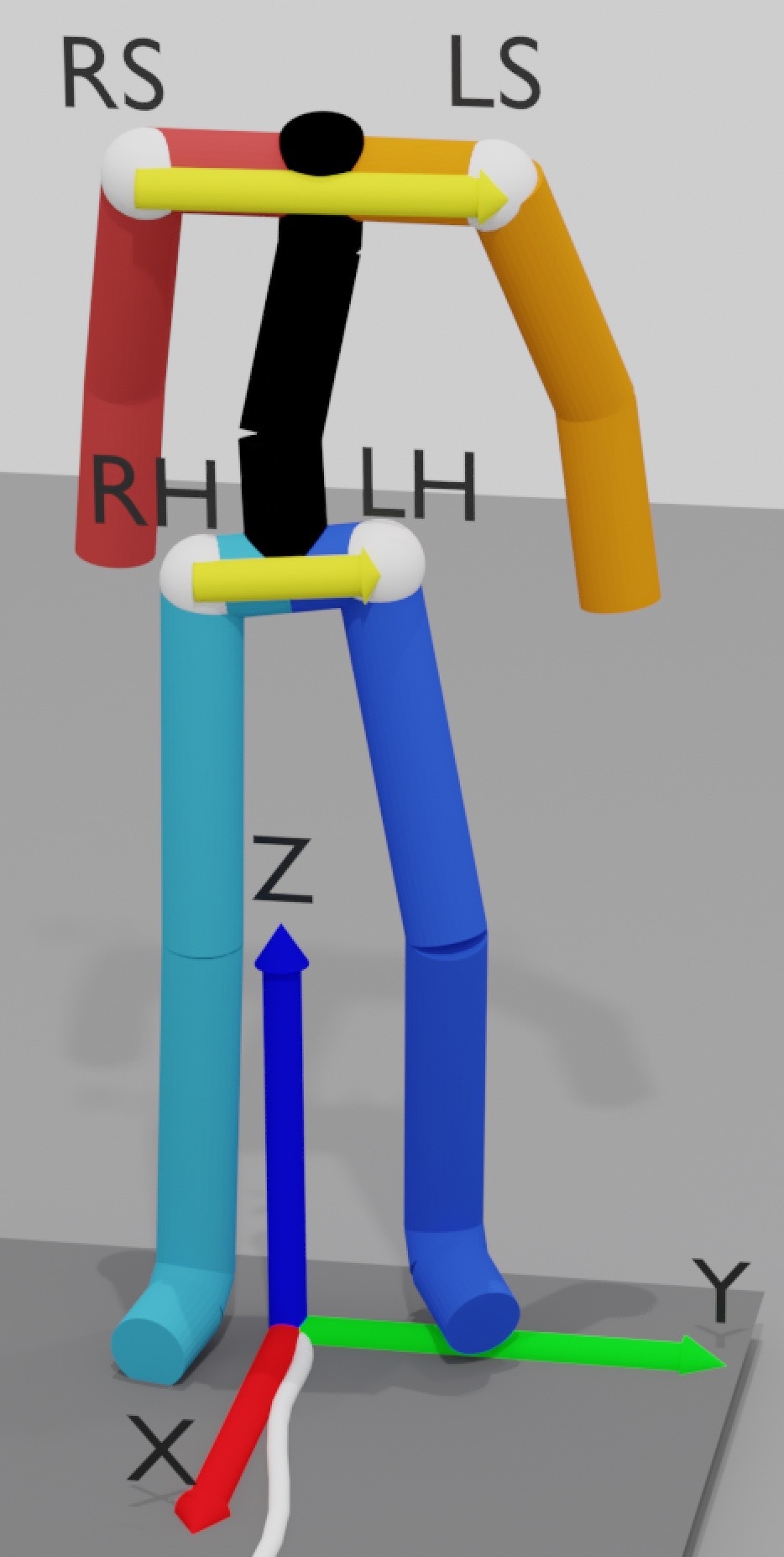}
    \caption{\textbf{Body's local coordinate system:} The origin is defined as the projection of the root joint into the ground.
    The X-axis direction is computed by taking the cross product between the average of the two yellow vectors: (left hip (LH) - right hip (RH)) and (left shoulder (LS) - right shoulder (RS)). The Z-axis (gravity axis) remains the same, and the Y-axis is the cross product of Z and X. }
    \label{fig:skeleton}
\end{figure}

\section{Motion representation}
\label{app:sec:representation}

Here, we describe in detail the two motion representations
employed in this work: skeleton-based (Section~\ref{subsec:skeleton}) and SMPL-based (Section~\ref{subsec:smpl_repr}). We standardize both of them by subtracting the mean and dividing with the standard deviation across the training data.

\subsection{Skeleton-based representation}
\label{subsec:skeleton}

We employ the representation introduced by Holden~et.~al.~\cite{holden2016motionsynthesis}. As input, we use the 21 joints from the Master Motor Map (MMM) framework~\cite{mmm_terlemez}.
For each frame, we encode: \begin{enumerate}[label=\roman*)]
\item the \textbf{3D human joints} in a coordinate system local to the body (the definition of the 3 axes and the origin is explained in Figure~\ref{fig:skeleton}) without the root joint (20x3=60-dimensional features), 
\item the \textbf{angles} between the local X-axis and the global X-axis (by storing them as differences between two frames) (1-dimensional feature),
\item the \textbf{translation}, as the velocity of the root joint in the body's local coordinate system for X and Y axes, and the position of the root joint for the Z axis (3-dimensional features).
\end{enumerate}

Then, we concatenate this into a feature vector in $\mathbb{R}^{64}$. For a motion sequence of duration $F$, the data sample will be in $\mathbb{R}^{F \times 64}$.

Note that, as we store the difference of angles for ii), when we integrate the angles, we assume that the first angle is 0 (which means that the body's local coordinate system is aligned with the global one). This is a way of canonicalizing the data so that each sequence starts with the body oriented in the same way. 

\subsection{SMPL-based representation}
\label{subsec:smpl_repr}

To construct the feature vector for SMPL data, we store:
\begin{enumerate}[label=\roman*)]
\item the \textbf{SMPL local rotations}~\cite{smpl2015} (parent-relative joint rotation according to the kinematic tree) in the 6D continuous~\cite{Zhou2019OnTC} representation (we load the AMASS data processed with SMPL-H~\cite{MANO:SIGGRAPHASIA:2017}, but we remove all rotations on the hands, resulting in 21 rotations) (21x6=126-dimensional features),
\item the \textbf{global rotation} in the 6D continuous~\cite{Zhou2019OnTC} representation (6-dimensional features),
\item the \textbf{translation}, as the velocity of the root joint in the global coordinate system for the X and Y axes, and the position of the root joint for the Z axis (3-dimensional features).
\end{enumerate}

Then, we concatenate this into a feature vector in $\mathbb{R}^{135}$. For a motion sequence of duration $F$, the data sample will be in $\mathbb{R}^{F \times 135}$.

To help the network learn the global rotation, we canonicalize the global orientation (by rotating all bodies and trajectories) so that the first frame is oriented in the same direction for all sequences.

\section{Evaluation details}
\label{app:sec:evaluation_metrics}
In this section, we give details on the evaluation metrics %
and explain our implementation. We further provide
details on our human study.

\noindent\textbf{Metrics definitions.}
As explained in
\if\sepappendix1{Section~4.1}
\else{Section~\ref{subsec:datasets}}
\fi
of the main paper,
the metrics are not computed with the original 3D joint coordinates. 
All motions are first canonicalized (so that the forward direction of the body faces the same direction for all methods and the ground truth). Except for the root joint, all the other joints are expressed in the body's local coordinate system (see Figure~\ref{fig:skeleton}). Then, the APE and AVE metrics are computed separately on these transformed coordinates.

As in JL2P~\cite{Ahuja2019Language2PoseNL} and Ghosh et al.~\cite{Ghosh_2021_ICCV}, we define the evaluation metrics as follows, and compute them on the test set.
The Average Position Error (APE) for a joint $j$ is the average of the L2 distances between the generated and
ground-truth joint positions over the time frames ($F$) and test samples ($N$): %
\begin{equation}
\operatorname{APE}[j]=\frac{1}{N F} \sum_{n \in N} \sum_{f \in F}\left\|H_{f}[j]-\hat{H}_{f}[j]\right\|_{2} .
\end{equation}
We omit denoting the iterator $n$ (over the test set) from the pose $H$ for simplicity.
The Average Variance Error (AVE), introduced in Ghosh et al.~\cite{Ghosh_2021_ICCV} captures the difference of variations. It is defined as the average of the L2 distances between the generated
and ground truth variances for the joint $j$:
\begin{equation}
\operatorname{AVE}[j]=\frac{1}{N} \sum_{n \in N}\|\sigma[j]-\hat{\sigma}[j]\|_{2} 
\end{equation}
where,
\begin{equation}
\sigma[j]=\frac{1}{F-1} \sum_{f \in F}\left(H_{f}[j]-\widetilde{H}_{f}[j]\right)^2 \in \mathbb{R}
^3 \end{equation}
denotes the variance of the joint $j$. $\widetilde{H}[j]$ is the mean of the joint throughout the motion.

We report in the tables: \begin{enumerate}[label=\roman*), topsep=0pt]
\item the \textit{root joint} errors by taking the 3 coordinates of the root joint,
\item the \textit{global trajectory} errors by taking only the X and Y coordinates of the root joint (it is the white trajectory on the ground in the visualizations),
\item the \textit{mean local} errors by averaging the joint errors in the body's local coordinate system,
\item the \textit{mean global} errors by averaging the joint errors in the global coordinate system.
\end{enumerate}

\noindent\textbf{Metrics implementation.}
We were unable to reuse the evaluation code of Ghosh~et~al~\cite{Ghosh_2021_ICCV}
for two reasons: (i)~the code did not reproduce the results in \cite{Ahuja2019Language2PoseNL,Ghosh_2021_ICCV},
(ii)~there is a bug in the evaluation script \href{https://github.com/anindita127/Complextext2animation/blob/94c53c504062d970e891dfee4b3e5e76a7e3f079/src/eval_APE.py#L249}{\texttt{src/eval\_APE.py}} line 249, the slicing for the trajectory loss is wrong (notably \texttt{y[:, 0]} instead of \texttt{y[:, :, 0]}; this issue was confirmed by the authors). 

Furthermore, the function \texttt{fke2rifke} might also have slicing bugs (the implementation is the same in the codes of  \href{https://github.com/chahuja/language2pose/blob/a65d6857d504b5c7cc154260ee946224d387da9d/src/data/data.py#L145}{\texttt{JL2P}} and \href{https://github.com/anindita127/Complextext2animation/blob/d89730c17580c5b9ccb767d55257f6781eec062a/src/data.py#L161}{\texttt{Ghosh et al.}}) and is used indirectly for evaluation metrics. 

Finally, the JL2P~\cite{Ahuja2019Language2PoseNL} code release does not include the evaluation. 
We therefore chose to reimplement our evaluation which we compute directly
on 3D coordinates, rather than the standardized (mean subtraction, division by standard deviation) version as explained in
\if\sepappendix1{Section~4.1}
\else{Section~\ref{subsec:datasets}}
\fi
of the main paper.

\noindent\textbf{Perceptual study details.}
The results of the pairwise comparisons in
\if\sepappendix1{Figure~3}
\else{Figure~\ref{fig:humanstudy}}
\fi
of the main paper are obtained as follows.
We randomly sample 100 test descriptions and generate motion visualizations
(as in the supplemental video~\cite{projectpage_temos})
from all the three previous methods \cite{Ahuja2019Language2PoseNL,Ghosh_2021_ICCV,lin2018},
our method, and the ground truth (500 videos). From these visualizations,
we create pairs of videos for each comparison, randomly swapping
the left-right order of the video in each question (also 500 videos). %
For the semantic study, we display the text as well as the pair of videos simultaneously to Amazon Mechanical Turker (AMT) workers. The workers are asked to answer the question: ``Which motion corresponds
better to the textual description?''. For the realism study, we use the same set of motion pairs and display them to other AMT workers but without showing the text description. They are asked to answer the question: ``Which motion is more realistic?''.
For both studies, each worker answers a batch of questions,
where the first 3 questions are discarded and used as a `warmup' for the task.
We further added 2 `catch trials' to detect unqualified workers, whose batch we discarded
in our evaluation. We detected exactly 20 such workers out of 100 in both studies. Each pair of videos
is shown to multiple workers between 2 and 5 (4 in average), from which we compute
a majority vote to determine which generation is better than the other. If there is a
tie, we assign a 0.5 equal score to both methods. The resulting
percentage is computed over the 100 test descriptions.

For the semantic study, posing the question as a pairwise comparison can lead to a bias towards a more realistic
motion (although the semantic correspondence to the text may be worse).
To disentangle such realism 
bias in pairwise comparisons,
we repeat the semantic perceptual study with one video at a time: workers on AMT were asked to rate how much they agree with the following statement: ``\textit{The body motion correctly represents the textual description}'' by choosing between (1) \textit{Strongly disagree}, (2) \textit{Disagree}, (3) \textit{Neither agree nor disagree}, (4) \textit{Agree}, or (5) \textit{Strongly agree}.
Using a 20-sequence test set, we generated 4 batches with our method and 1 batch with Ghosh et al.~\cite{Ghosh_2021_ICCV}. At the beginning, we added 2 examples along with instructions to explain what we expect from `correctness’ and to disentangle from the realism: (i) a realistic motion (from ground truth) that does not correspond to the text, and (ii) a relatively 
less realistic motion (from generations) that does correspond to the text.
We added 4 warm-up examples after them, and 3 catch trials at random locations.
We ran the study with 24 AMT workers but half of them failed the catch trials (and the remaining results were not very consistent).
So we repeated this experiment with naive lab members. Again we used all the examples at the beginning and catch trials. With this setup, we obtained 21 completed batches.
The users rated an average of \textbf{3.50} correctness score out of 5 likert scale for our
generations versus \textbf{3.04} for Ghosh~et~al.~\cite{Ghosh_2021_ICCV}.
To further assess the quality of the diverse generations,
we generated 5 samples per test description for \methodshort{}
and obtained an average of \textbf{3.54 $\pm$ 0.1} score, showing
that we preserve correctness within diverse generations as well.

\section{Implementation details}
\label{app:sec:implementation}

\noindent\textbf{Architectural details.}
For all the encoders and the decoder of \methodshort{},
we set the embedding dimensionality to 256, the number of layers to 6, the number of heads in multi-head attention to 6, the dropout rate to 0.1, and the dimension of the intermediate feedforward network to 1024 in the Transformers.

\noindent\textbf{Library credits.}
Our models are implemented with PyTorch~\cite{NEURIPS2019pytorch} and PyTorch Lightning~\cite{falcon2019pytorch}. We use Hydra~\cite{Yadan2019Hydra} to handle configurations.
For the text models, we use the \texttt{Transformers} library~\cite{wolf-etal-2020-transformers}.

\noindent\textbf{Runtime.} Training our \methodshort{} model takes about 4.5 hours for 1K epochs, with a batch size of 32, on a single Tesla V100 GPU (16GB) using about 15GB GPU memory for training (i.e., \textit{16 seconds} per epoch).
In comparison, according to their paper, Ghosh~et~al.~\cite{Ghosh_2021_ICCV} trained their model for 350 epochs on a single Tesla V100 GPU in about 15 hours (i.e., \textit{154 seconds} per epoch). 
While the rest of the hardware specifications or implementation efficiency may vary, given the same type of GPU for both methods, we can expect that our model trains an order of magnitude faster. 
This may be because our model generates the full motion with only one decoder pass.
Previous work produces one frame at a time iteratively (i.e., the next frame has to wait for the previous one to be generated).

\section{KIT Motion-Language text statistics}
\label{app:sec:statistics}

In the KIT Motion-Language~\cite{Plappert2016_KIT_ML} dataset, there are 3911 motions and a total of 6352 text sequences (in which 900 motions are not annotated).
Using a natural language processing parser, we extract ``action phrases'' from each sentence, based on verbs. For example, given the sentence ``A human walks slowly'', we automatically detect and lemmatize the verb, and attach complements to it, such that it becomes ``walk slowly''. With this procedure, we group sequences that correspond to the same action phrase and detect 4153 such action clusters out of 6352 sequences. The distribution of these clusters is very unbalanced: ``walk forward'' appears 596 times while there are 4030 actions that appear less than 10 times (3226 of them appear only once). On average, an action phrase appears 2.25 times. This information shows that the calculation of distribution-based metrics, such as FID, is not relevant for this dataset.

\end{document}